\documentclass[10pt]{article}
\usepackage[utf8]{inputenc}

\usepackage{natbib}
\usepackage{enumitem}

\usepackage{setspace}

\usepackage[final]{pdfpages}
\usepackage{amsmath}
\usepackage{graphicx}

\usepackage{hyperref,color}

\usepackage{alltt}
\usepackage{xcolor}

\usepackage{array}
\usepackage{amssymb}
\usepackage{extarrows}
\usepackage{graphicx}
\usepackage{subcaption}

\usepackage{lineno} 

\usepackage{mathtools}

\usepackage{float}
\usepackage[section]{placeins}

\modulolinenumbers[5]

\usepackage{xcolor}

\usepackage{sectsty}

\sectionfont{\Large}
\subsectionfont{\large}

\usepackage{amsthm}
\theoremstyle{plain}

\theoremstyle{definition}

\title{Abstraction, Reasoning and Deep Learning: A Study of the ``Look and Say" Sequence}

\author{Wlodek W. Zadrozny$^{1,2,}$ \\
wzadrozn@uncc.edu\\
$^1$ College of Computing, University of North Carolina at Charlotte\\
$^2$ School of Data Science, University of North Carolina at Charlotte}

\date{version 2.0 January 2022}

\begin{document}
\maketitle
\begin{abstract}

{The ability to abstract, count, and use System~2 reasoning are well-known manifestations of intelligence and understanding. In this paper, we argue, using the example of the ``Look and Say" puzzle, that although deep neural networks can exhibit high `competence' (as measured by accuracy) when trained on large data sets (2 million examples in our case), they do not show any sign on the deeper understanding of the problem, or what D. Dennett calls `comprehension'. 
We report on two sets experiments: first, computing the next element of the sequence, and ,then, the previous element. We view both problems as building a translator from one set of tokens to another.  We apply both standard LSTMs and Transformer/Attention-- based neural networks, using publicly available machine translation software. We observe that despite the amazing accuracy, the performance of the trained programs on the actual L\&S sequence is bad, and shows no understanding of the principles behind the sequences. 
The ramifications of this finding include: (1) from the cognitive science perspective, we argue that we need better mathematical models of abstraction; (2) the universality of neural networks should be re-examined for functions acting on discrete data sets.; (3) we hypothesize topology  can provide a definition of without the reference to the concept of distance.
}

\end{abstract}


\section{Introduction}

The ``Look and Say" (L\&S) sequence is often given as a puzzle: 
``What is the next number in the sequence 1, 11, 21, 1211, 111221?"
The pattern is discovered by noting the counts of digits. Thus 
\begin{quote}
1 counts as ``one 1" and generates 11.

11 counts as ``two 1s" and generates  21.

21 counts as ``one 2, then one 1" and generates  1211.

1211 counts as ``one 1, one 2, then two 1s" and generates  111221.

111221 counts as ``three 1s, two 2s, then one 1" and generates  312211.
\end{quote}
\noindent
The pattern can be challenging to discover by humans, but, in the author's experience, most people get it after seeing half a dozen examples; saying the sequence out loud seems to help. Likely, the challenge for humans comes from having to mix the parsing of a number into digits, counting each sequence of identical digits, and then generating the next number. 
In addition, the puzzle is different than typical examples of number sequences given at school.

A few more words about the pattern. The sequence grows very rapidly, and reaches the length of 12,680,852 digits after 59 iterations, with each subsequent number being about 30\% longer that the previous 
one;\footnote{\url{https://archive.lib.msu.edu/crcmath/math/math/l/l407.htm}} even at 25 steps we get a relatively long sequence.\footnote{ \url{https://oeis.org/A005150/b005150.txt}}  
The On-Line Encyclopedia of Integer Sequences\footnote{OEIS: \url{https://oeis.org/A005150}} has additional information about the sequence and the main references. 
Given the simplicity of the puzzle, we can ask how good deep neural networks are at this task. 

\begin{figure}[]
    \centering
\includegraphics[width=0.49\columnwidth]{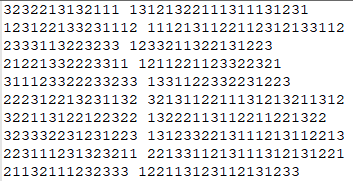}
\caption{Example training data. The left column shows the input and the right column the expected value, according to the ``Look and Say" rule discussed in this article.  }\label{fig:trnData}
\end{figure}

\textit{Why is this pattern interesting for machine learning?} 
Per Stanovich and West \cite{stanovich2000individual}, and Kahneman \cite{kahneman2011thinking}, we can model cognition using the metaphor of two systems.
System 1 operates automatically and intuitively, while 
System 2 is about controlled processing, e.g. deliberation, reasoning, and computing. 

The L\&S pattern requires a degree of abstraction, and clearly goes beyond developing a simple association (which is a basis of System 1). Moreover, failures of machine learning (ML) programs on L\&S,  unlike e.g. in textual entailment, cannot be attributed to the lack of commonsense knowledge or failures of NLP. \\

\textit{This article is organized as follows.} 
Since we are not aware of any machine learning work on L\&S, in the next few sections we describe our experiments and observe that, even though the overall accuracy of the ML programs is very high, the systems clearly have no clue what they are doing. Section \ref{sec:Methods} covers the experiments with the L\&S sequence, that is, computing the next element of the sequence; and then Section \ref{sec:rLS} is about computing the previous element.

Based on these simple experiments, in Section \ref{sec:discussion}, we engage in a more general discussion of neural networks.
First, we observe that neural networks (NNs) do not perform well on tasks requiring a degree of abstraction, as different as humor and algebra. Then, contrary to the common perception, we note that NNs are \textbf{not} universal approximators. The often-cited theorems only pertain to the approximations of continuous functions. However, some problems such as L\&S or integer addition are naturally  defined on discrete data points, that is, as functions from sequences of digits to sequences of digits. When formulated as learning problems, we realize that 
 there are no canonical continuous approximations of such functions. Therefore, the poor performance of deep learners could perhaps be traced to this indeterminacy. If this were to be  the case, ideas from topology and topological data analysis (TDA) might help provide a solution, because in topology, continuity can be defined without the recourse to the notion of distance. Also, TDA often has a measurable impact when combined with deep learning. The final Section summarizes all of this and speculates about possible connections to other machine learning problems, and in particular to ``few-shot learning."


\section{L\&S Data}\label{sec:LSdata}

In principle, we can easily generate the data, however, there are two practical problems. First, neural networks require lots of data, and the length of the L\&S sequence grows exponentially, and therefore only a small initial segment would be easily representable as a standard data structure. But if we reformulate the problem as learning to group and count, there is abundance of data.

As mentioned above, we trained on 2 million examples. The examples, like in Figure~\ref{fig:trnData}, are taken from sequences of digits ``1", ``2", and ``3," up to 15 digits long (for input). This was optimal for being able to complete an experiment on Google Colab in 12 hours or less. 

The training data can be easily obtained. 
There are 3,486,784,401 unique strings of length 20 generated from the alphabet ``1 2 3"; 129,140,163 of length 17; 14,348,907 of length 15; and 59,049 of length 10. 

We need to limit ourselves to sequences containing the runs of length at most 3. The file used 
in experiments was obtained by replacing all runs of length 4 or more by runs of length 3 of the same digits. Thus the number of elements in the file is still 14,348,907, but there are many repetitions. A small sample of the training data is shown in Figure~\ref{fig:trnData}. The test data has the same format. 
From this file we obtained a random sample of train data and test data. The train data was the same in all experiments and consisted of 2M pairs, but we typically chose a sample of 1K to 10K items for testing.

\section{Methods}\label{sec:Methods}

Regarding the experimental setup.  In principle, we could use any neural network for which there exists a proof of the universality property, i.e. ability to approximate any function to any degree of accuracy.\footnote{Wikipedia has a good introduction \url{https://en.wikipedia.org/wiki/Universal_approximation_theorem}} However,  in practice, researchers build complicated networks for specific purposes. 
Therefore, we settle on a plausible setting where we view the problem as machine translation from one sequence of tokens to another.

In this section, we report on two series of experiments, both of which view the problem as machine translation. This is a natural metaphor: we have a sequence of tokens, and we try to generate a new sequence. The first set of experiments was first performed in 2018 (and repeated for this report) used LSTMs, and in the second one, from the summer of 2021, we used attention. In both cases, we used publicly available  programs. 
We trained on a very large set of examples (2 million) and tested on up to 20 thousand examples. \\

\begin{figure}[h]
    \centering
\includegraphics[width=0.99\columnwidth]{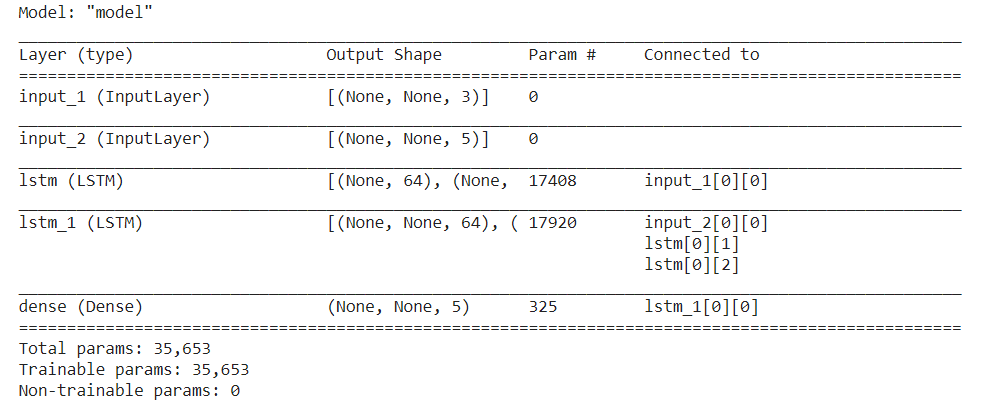}
\caption{A standard LSTM-based translation model used for an experiment with the ``Look and Say" puzzle.
The results on the L\&S sequence are shown in Figure \ref{fig:lstmResults}.
}    \label{fig:lstmCholet}
\end{figure}

We believe the experiments show that high accuracy does not imply understanding. 
We ran the experiments in dozens of different settings/hyperparameters, 
(all resulting in the same conclusions). Here, we describe the one where we performed the most systematic evaluations.

\subsection{L\&S problem as translation without attention }\label{sec:lstmNoAtt}

In this experiment we used a standard, LSTM-based model. The results discussed below are for the one 
depicted in Figure~\ref{fig:lstmCholet}.\footnote{We experimented with several versions of the code. 
The validation losses varied between 5\% and 1.5\%. The code was a modification of \url{https://github.com/keras-team/keras/blob/master/examples/lstm_seq2seq.py}, downloaded a few years ago.}

\begin{figure}[]
    \centering
\includegraphics[width=0.8\columnwidth]{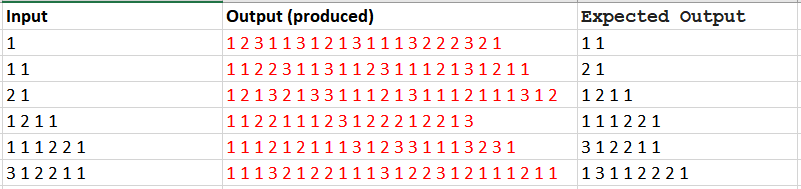}
\caption{A standard LSTM-based translation model (see the architecture in Fig. \ref{fig:lstmCholet}) clearly does not `understand' the ``Look and Say" puzzle despite training on 2M examples, and a very low loss of 1.5\%.}\label{fig:lstmResults}
\end{figure}


The original code was used to demonstrate machine translation using LSTMs and word embeddings. The choice of 64 dimensions in this experiment was based on the work of \cite{patel2017towards} (and for higher dimensions the results were worse). 

The system was trained on  2,000,000 (2M) samples of unique pairs where the input sequence length was up to 15 digits.
The hardware was Tesla P100 running on Google Colab Pro. It took about 45 minutes to train. The validation loss was about 1.5\% and stable ($\pm$ 0.07\%) after 92 epochs. The batch size was 512.

The system clearly has not learned (from 2M examples!) the pattern of the puzzle, as shown in its output shown in Figure \ref{fig:lstmResults}.

\subsection{L\&S problem as translation, using attention}\label{sec:lstmAtt}

In the next series of experiments, again, we viewed L\&S as a translation problem. We used another standard piece of code, this time from the Google Colab library (June 2021 version of the tutorial "Neural machine translation with attention" 
({\small{\url{https://colab.research.google.com/github/tensorflow/text/blob/master/docs/tutorials/nmt_with_attention.ipynb}}}).
We did not change the preprocessing nor any parameters except for the vocabulary size (500 in the second experiment, 12 in the third, and 7 in the fourth).

The data, batch size and hardware (Tesla P100) were the same as in the previous experiment. In both cases, we set the number of units to 1024, and the number of epochs to 10 (each epoch taking about 45 minutes of compute time). We tried vector dimensions of 256; and then 64, and finally 128.
We clearly saw the improvement in accuracy (compared to the model without attention) -- after half an epoch, we got the previous 1.5\% loss (batch loss in one of the experiments).
The final batch losses were  1.2765e-04, 1.5548e-04, and 1.0073e-04.

\begin{figure}[t]
    \centering
\includegraphics[width=1.0\columnwidth]{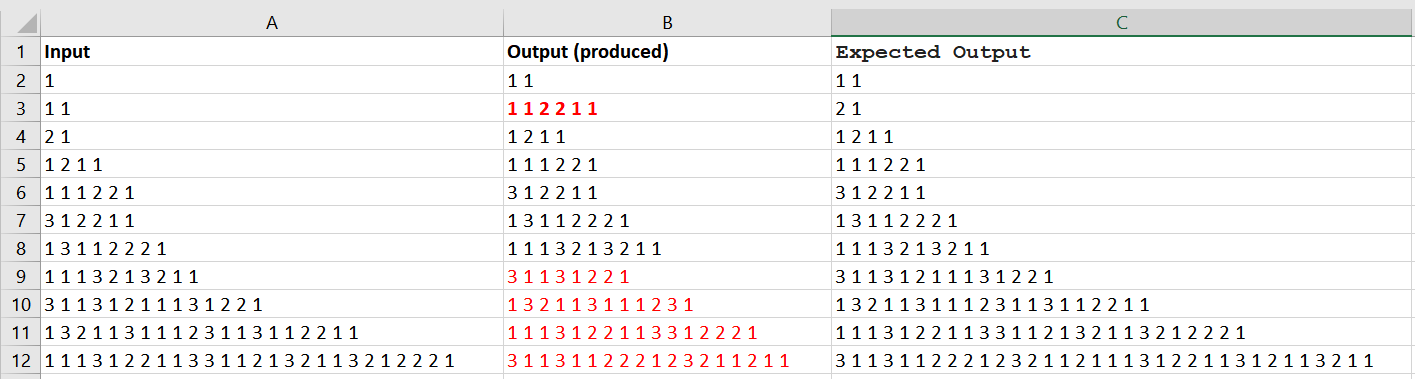}
\includegraphics[width=1.0\columnwidth]{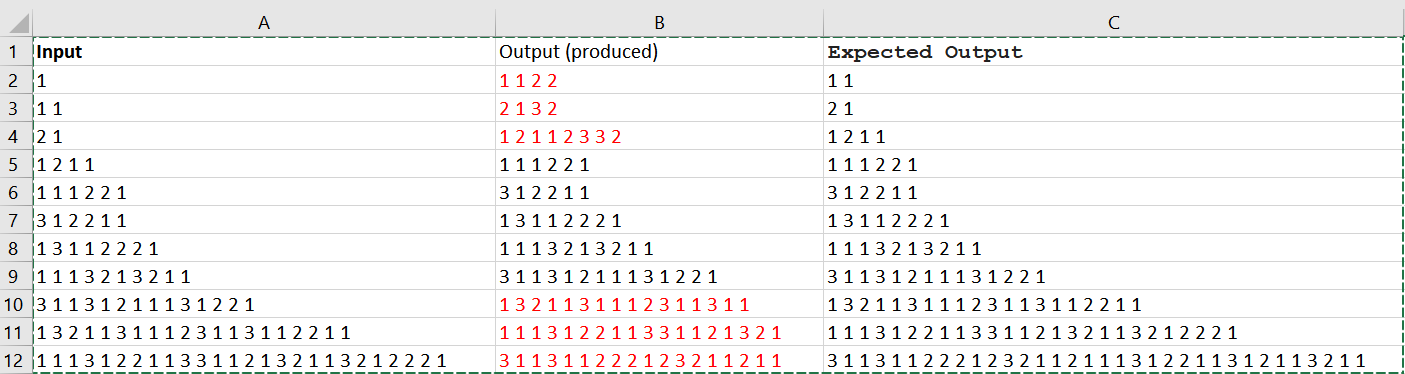}
\caption{Despite very high overall accuracy, our half a dozen experiments with attention-based models show that using transformers does not produce understanding. This figure shows two examples of the typical mistakes (in red). In our experiments, the test data error rate varied between 33 and 158 on a sample of 10,000.}
\label{fig:lstmExpAtt}
\end{figure}

\subsection{Reversed L\&S Sequence}\label{sec:rLS}

Given the above experiments, one can ask whether the disappointing performance of deep neural networks (DNNs) is due to the fact that the L\&S sequence is fast growing. Therefore, we can ask what happens when we reverse the order and try to learn the Reversed L\&S sequence. 
That is generating 1 from 11, 21 from 1211 etc. 
Note this is a simpler problem: all odd numbered digits specify the number of repetitions that should be generated for the even numbered digits. That is `312211' means generate three 1's, two 2's and one 1. 

The data was the same as in Section \ref{sec:LSdata}. 
We used the newer (Jan 2022) version of the TensorFlow tutorial
``Neural machine translation with 
attention (\small{\url{https://www.tensorflow.org/text/tutorials/nmt_with_attention}})." 
\normalsize The only parameters changed were:  
 \texttt{batch\_size} -- to 128 from  64, \texttt{max\_vocab\_size} to 12 from 5000,  and the dimension of embeddings \texttt{embedding\_dim} -- we used both 64 and 128 (in different experiments). The number of units remained 1024. We ran several experiments, up to 14 epochs, and in all of them the batch accuracy seems to stabilize between $10^{-4}$ and $10^{-5}$ after four to six iterations.


\section{Results}\label{sec:results}

\noindent
\textit{Did the transformers solve the L\&S puzzle?}
As shown in Figure \ref{fig:lstmExpAtt},   
the programs arguably had no clue how to solve the puzzle, despite their amazing accuracy. (In one experiment, testing on 10,000 examples produced only 33 errors!)
Moreover, the pattern of the solution i.e. `count' and `indicate the digits' could be applied to any number, e.g. 
``446988" will produce ``24161928". However, in other experiments, not reported here, we observe the same lack of understanding on arbitrary sequences of digits. 

The low batch losses described in Section~\ref{sec:Methods} did not translate into higher test accuracy --- in the last case there were 158 errors in a sample of 10,000 examples.
Also, notice that should a network get the rule behind L\&S in any batch, all the subsequent batches would have a perfect accuracy. Obviously, this did not happen. \\

\begin{figure}[h]
    \centering
\includegraphics[width=0.8\columnwidth]{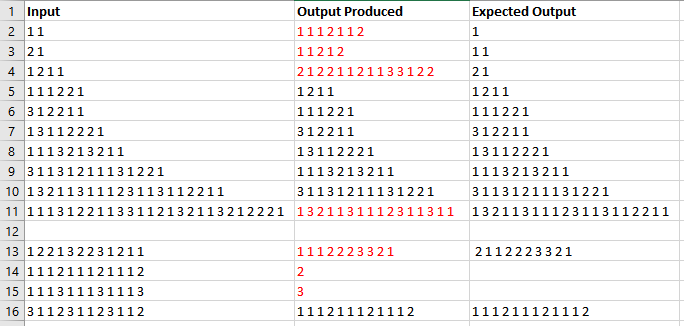}
\caption{The attention-based translation model  clearly does not `understand' the ``Reversed Look and Say" puzzle despite training on 2M examples, and a very low batch loss of  9.2713e-05. Errors are shown in red.}\label{fig:RgMTResults}
\end{figure}

\noindent
\textit{Did the transformers solve the reversed L\&S puzzle?}
Even for this simplified problem, the transformer-based program shows clear lack of understanding. Despite very high accuracy, which seems higher than on the L\&S data, as  we typically see only 10--20 errors on 20 thousand test examples, the errors shown in 
Figure~\ref{fig:RgMTResults} are typical. Clearly, they would be impossible with even elementary understanding of what this problem is about.  Notice that lines 13-16 are all of the same length but only one of the four examples is correct. 

Thus, as with the original L\&S puzzle, the reversed L\&S sequence is a challenge for a standard transformer-based program.

\section{Discussion and Related Work}\label{sec:discussion}

The key question the reader may ask is: Why should anyone care about these puzzles? 
--- 
We already suggested an answer to the first question in the Introduction, namely the L\&S test is about System 2 thinking 
(\cite{stanovich2000individual},  \cite{kahneman2011thinking}), and therefore 
raises the question whether deep NNs can properly model System 2 thinking (Section \ref{sec:cogAsp}).

The second reason has to do with 
 the puzzle touching upon some fundamental issues in deep learning,
 namely, for which classes of functions  LSTMs, attention-based models, and, generally, deep neural networks are universal approximators (Section \ref{sec:mathAsp}). In Section \ref{sec:discon}, we will speculate that for some problems, L\&S included, we might need to broaden the notion of continuity to account for situations when there is no unique metric gradient.

\subsection{Cognitive aspect: Humor and algebra -- why can't NNs get it?}\label{sec:cogAsp}

Why should we put humor and algebra together in a question about NNs abilities? They both require a degree of abstraction. Like L\&S, humor requires abstracting from two qualities; per 
\cite{ginzburg2015understanding}, 
``laughter is an event anaphor that can convey at least two distinct meanings: the
enjoyment of an event and the recognition of an incongruous event."
In algebra, obviously, multiple abstractions are abundant. 
Yet, as observed by \cite{mitchell2021abstraction}, ``no current AI system is anywhere
close to a capability of forming humanlike abstractions or analogies." The point of playing with L\&S -- like puzzles is to create closed systems on which experiments are easier to perform. We  view the L\&S puzzle as requiring both abstraction and combining two different operations.

Another issue is distinguishing intelligence from competence. D. Dennett in \cite{millhouse2021foundations} 
argues 
that animals ``display competence with tasks whose purpose and rationale
they do not comprehend" while a human would have ``the capacity to understand what they are doing and why." The performance of our models on the puzzle seems to illustrate this distinction very well. 
Namely, the overall accuracy is high, but the lack of understanding of the problem is very obvious from Figures \ref{fig:lstmResults} and \ref{fig:lstmExpAtt}. In this context, it's worth mentioning the results of 
\cite{talmor2020leap} showing that the accuracy of pretrained models is low on tasks that combine counting with other knowledge skills. 
We can think of the training on 2M examples as the creation of a pretrained model, and then testing on the initial segment of 
the L\&S puzzle as task combining counting with another knowledge skill (grouping).\\

We see it as an open question, whether focusing on the deliberative cognitive System 2  is needed to achieve progress in AI/Deep Learning. 
A large portion of the recent work in the System 2 space 
focuses on inference, rule induction and compositionality (e.g. 
\cite{clark2020transformers}, \cite{richardson2020probing},\cite{yanaka2021sygns},  
\cite{nye2021improving},
\cite{lake2018generalization},
\cite{lake2017building},\cite{nye2021improving})
Some researchers postulate a program of explicitly connecting AI with Cognitive Science (e.g. \cite{MarcusDavis2021}).
The L\&S puzzle is about forming abstractions based on grouping and counting, and complements the work done on these other aspects.

\subsection{Mathematical aspect: Are NNs universal approximators?}\label{sec:mathAsp}

Intuitively, the process of abstracting requires a change of representation, and therefore abstractions introduce conceptual `discontinuities.' We are not sure how to think about abstraction mathematically, but a lot is known about continuity. 

From a mathematical perspective, the two L\&S puzzles go to the core of NNs capacities. In theory, NNs can approximate `any' function. The actual theorems 
(\cite{cybenko1992approximation},\cite{hornik1991approximation},
\cite{yun2020transformers})  put  restrictions, e.g. the learned functions must be continuous and differentiable. Thus, if System 2 reasoning requires representation of `discontinuities,' does it mean that some problems cannot be solved by neural networks? 

The L\&S function can be viewed as continuous if its domain is  extended to real numbers, where both input and output values are contained. It is easy to see that there are infinitely many  continuous and differentiable functions from reals to reals that agree with the L\&S function (same for Reversed L\&S). Therefore, the universality of NNs should apply. Yet, there are practical and theoretical obstacles here. 

On the empirical side, a recent publication on machine learning of mathematical structures  \cite{he2021machine} observes that there seem to be a hierarchy of mathematical problems ordered by their amenability to use machine learning (ML) methods, with numerical analysis being the easiest and combinatorics and analytic number theory being the hardest. Note that the latter two deal with properties of discrete structures, even if they apply methods of mathematical analysis. In a related work,
 \cite{lample2019deep} are surprised by ``the difficulty of neural models to perform simpler tasks like integer
addition or multiplication." Given that they ``can be applied to difficult tasks like function integration, or solving differential equations". 

There are also some interesting results about the deep learning models used in our experiments. Thus, \cite{hahn2020theoretical} shows that self-attention ``cannot model periodic finite-state languages, nor hierarchical structure, unless the number of layers or heads increases with input length." 
In particular, this means that evaluating the parity of arbitrary expressions cannot be reliably evaluated by transformers --- as we observed in Section \ref{sec:rLS}  Reversed L\&S touches on a parity aspect.  
On the other hand, \cite{yun2020transformers}
show that transformer models ``can universally approximate arbitrary continuous sequence-to-sequence functions on a compact domain." In another set of results, 
\cite{bhattamishra2020computational} prove several results showing transformers to be Turing-complete, while in \cite{bhattamishra2020ability} it is observed, that ``surprisingly, in contrast to LSTMs, transformers do
well only on a subset of regular languages." We are not sure how these two views can be productively reconciled.

Other recent work focusing on limitations of deep learning, noted but not discussed here, are \cite{dong2021attention} and \cite{yehuda2020s}. Finally, in \cite{zadrozny2020towards} we gathered evidence (from other publications) for improved performance of machine learning on many tasks related to natural language processing (NLP) when the outputs are constrained e.g. by rules. The data used in NLP is discrete (word embedding notwithstanding), and therein lies the connection to this work. 

\subsection{Should we be introducing topologies?}\label{sec:discon}

The L\&S task is perhaps more complex that the problem of evaluating the mentioned above parity of bitstrings or learning integer addition. Yet, the three problems have a few things in common and Reversed L\&S has a clear parity aspect. We see at least three things in common:
\begin{enumerate}[label=(\alph*)]
\item they operate in spaces of discrete points, if viewed as the tasks of learning a function on sequence of digits;
\item { there is no canonical continuous (and differentiable) function that would extend them to a compact metric space};
\item they require some counting (parity or addition-carry).
Intuitively their values depend on some `global' property of the strings and for parity oscillate between $0$ and $1$.
\end{enumerate}
  
Based on these intuitions, we would like to speculate that the 
machine learning of discontinuities could be accomplished using 
a conceptual framework that does not require the computation of numerical gradients. We would like to hypothesize that the answer might be in the mathematical discipline of topology.

Classical set theoretic topology is all about continuous transformations (e.g. \cite{munkres2000}, \cite{engelking1968outline}), 
where \textit{continuity} is defined \textit{without} the concept of  `distance.' Namely, it does it by declaring some sets to be open and defining continuous functions as those where the inverse images of open sets are open.  We can even define the concepts of continuity and differentiability for maps between collections of points (see \cite{saveliev2016topology} for an introduction). 

Regarding point (b), Wikipedia has an introductory article about finite topologies which, among other things, shows the exponential growth in the number of possible topologies for a set with $n$ points\footnote{\url{https://en.wikipedia.org/wiki/Finite_topological_space}}. Thus, without additional constraints, there is no canonical way of applying gradient based algorithms to these particular three problems (L\&S, parity and addition).

Another intuitive argument for the difficulty in applying gradient-based algorithms comes from a 
post\footnote{
{https://www.physicsforums.com/threads/gradient-vector-without-a-metric.958243/}} 
 on the Physics Forum: the state of an ideal gas can be characterized by its pressure and temperature $(P,T)$. If the state is changing as a function of time, we can meaningfully talk about the ``velocity" of the state  $ V^P = \frac{dP}{dt}$, $V^T = \frac{dT}{dt} $  
which is `a kind of vector, a tangent vector.' However, it makes no sense of talking about a metric $ \sqrt{(V^P)^2 + (V^T)^2}$  since the units of each coordinate are different. 
This suggests that the applicability of gradient-based NNs solution requires us to properly conceptualize both L\&S and RL\&S puzzles as strings combining different units. In other words, explicit hints, or a partial solution. 

As to point (c), the intuition that topologies might be helpful is reinforced by the fact that 
both the parity and the carry operation in addition are naturally connected to concepts from topology. Thus, the parity appears in the definition of the oriented simplex (using permutations) \cite{munkres2018elements}, or well-known Euler formula and its 
generalization.\footnote{\url{https://en.wikipedia.org/wiki/Euler_characteristic}}
 The carry operation has a natural representation as a `cohomology' \cite{isaksen2002cohomological} -- a very important concept in algebraic topology and in topological data analysis 
 (\cite{munkres2018elements}, \cite{edelsbrunner2010computational}). 
 
 An  explicit connection between topology, deep learning and formal languages is investigated in \cite{ackerman2021formal}. In the space of continuous models and adversarial deep learning examples, there are explicit connections to topology, e.g., ``the set of adversarial negatives is of extremely low probability, and thus is never (or rarely) observed in the test set, yet it is dense (much like the rational numbers)" \cite{szegedy2013intriguing}, and ``manifold-based defenses need to be aware of the topology of the underlying
data manifold" 
\cite{jang2020need}.
There is also empirical support for using topology (via Topological Data Analysis, i.e. TDA) as an additional set of features for machine learning tasks, including deep learning, e.g. \cite{hofer2017deep,guss2018characterizing,
hu2019topology,Moor20Topological,carlsson2020topological},
\cite{kushnareva2021} and our own work \cite{doshi2018slsp,savle2019topological,gholizadeh2020topological}.

We hypothesize that for the three problems (L\&S, parity, and integer addition), the reason for the unconvincing performance of deep learning, might lie in the indeterminacy of continuous extensions of the functions defined on discrete sets.

\section{Conclusion}
In this article we identified the ``Look and Say" puzzle as an interesting problem for machine learning. We described a few simple experiments, 
on the original, L\&S,  and simplified, reversed L\&S of the puzzle. 
We concluded that, despite very high accuracy on the test data, the programs (produced with and without the attention mechanism) do not solve the puzzles, as shown in Figures \ref{fig:lstmResults}, \ref{fig:lstmExpAtt}, and \ref{fig:RgMTResults}. Since they make simple errors on the initial segments of the puzzle sequence, and generally on the test data, we infer that they lack the understanding of the problem. We  attribute this lack of comprehension to the indeterminacy of continuous extensions of
functions from sequences of digits to sequences of digits on the L\&S datasets.

Another hypothesis expressed in this article is the intuitive 
connection between the puzzle and the ability to abstract. Thus,
finding a formalization which would allow us to look at abstraction as a type of discontinuity (\cite{mitchell2021abstraction} could be an example of motivating recent work in this space). 

One more thing about the L\&S pattern might be worth mentioning in this discussion. Given the fast growth of the sequence, the puzzle seems related to the machine learning subbranch of 
``few-shot learning" \citep{lake2011one, fei2006one,wang2020generalizing} (because of the data sparsity of the actual L\&S sequence). However, in contrast with the cited work, the task discussed in this paper is not categorization or inference, but a combinatorial use of multiple (two) faculties --- perceiving and counting. 

In summary, we presented results of experiments and a discussion of the ``Look and Say" puzzle, and its simpler variant, "Reversed Look and Say".
The puzzle is interesting because it touches on several areas of active machine learning research: machine inference, learning of mathematical structures, and the universality of  neural networks. We hypothesize that the reason for the unconvincing performance of the example deep learning models lies in the indeterminacy of continuous extensions of the functions defined on discrete sets.\\

\textbf{Acknowledgment.} The author would like to thank Michael Robinson and 
Gonzalo Polavieja for comments on the previous version \citep{zadrozny2021abstraction} of this article.


\vspace{6pt}

\noindent
\textbf{Notes: } WZ performed the experiments and wrote the manuscript.
{This research received no external funding.
The author declares no conflict of interest.\\

\noindent 
\textbf{Abbreviations}
The following abbreviations are used in this manuscript:

\noindent 
\begin{tabular}{@{}ll}
L\&S & Look and Say \\
NNs & Neural Networks\\
dX & differential (in calculus) \\
AI & Artificial Intelligence \\
LSTM & Long-Short Memory Network \\
\end{tabular}}
\small



\end{document}